\documentclass[10pt,twocolumn,letterpaper]{article}

\usepackage{cvpr}              % To produce the CAMERA-READY version
\usepackage{multirow}

\definecolor{cvprblue}{rgb}{0.21,0.49,0.74}
\usepackage[pagebackref,breaklinks,colorlinks,allcolors=cvprblue]{hyperref}

\def\paperID{*****} % *** Enter the Paper ID here
\def\confName{CVPR}
\def\confYear{2025}

\title{EF-VI: Enhancing End-Frame Injection for Video Inbetweening}

\author{
    Liuhan Chen\textsuperscript{\rm 1}\quad
    Xiaodong Cun\textsuperscript{\rm 2}\thanks{Corresponding authors}\quad
    Xiaoyu Li\textsuperscript{\rm 3}\quad
    Xianyi He\textsuperscript{\rm 1,4}\quad
    Shenghai Yuan\textsuperscript{\rm 1,4}\\ 
    \textbf{Jie Chen\textsuperscript{\rm 1}\quad
    Ying Shan\textsuperscript{\rm 3}\quad
    Li Yuan\textsuperscript{\rm 1}\footnotemark[1]}\\ \\
    \textsuperscript{\rm 1}~Shenzhen Graduate School, Peking University\quad \textsuperscript{\rm 2}~GVC Lab, Great Bay University\\
    \textsuperscript{\rm 3}~ARC Lab, Tencent PCG\quad
    \textsuperscript{\rm 4}~Rabbitpre Intelligence \\ \\
    \url{https://github.com/GVCLab/Sci-Fi}
}

\begin{document}
 \twocolumn[{
 \renewcommand\twocolumn[1][]{#1}%
 \maketitle
 \label{pipe_comparison}
 \begin{center}
     % \captionsetup{type=figure}
     % \vspace{-4.5\baselineskip}
     \includegraphics[width=1.0\textwidth]{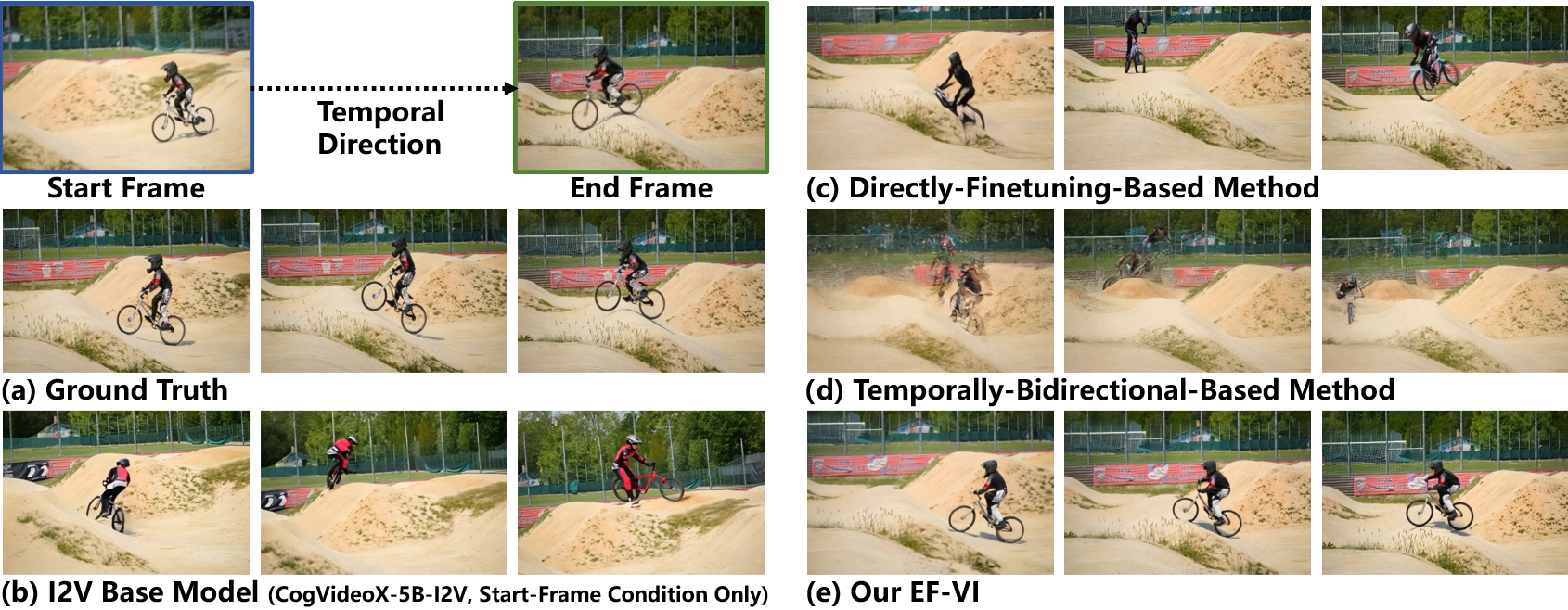}
     \captionof{figure}{
     \label{pipe_comparison}
     \textbf{Comparison between current Image-to-Video Diffusion Models (I2V-DM)-based methods and our EF-VI.} All the methods are based on the same I2V-DM, \emph{i.e.}, CogVideoX-5B-I2V~\cite{yang2025cogvideox}.
     According to the results, our method can produce more harmonious transitions in the challenging scenario.
 (a)~Ground truth.
 (b)~I2V results (conditioned on the start frame only) of the base model.
 (c)~Video inbetweening results of the directly-finetuning-based method.
 (d)~Video inbetweening results of the temporally-bidirectional-based method.
 (e)~Video inbetweening results of our EF-VI.
     }
 \end{center}
 }]
\begin{abstract}
Video inbetweening aims to synthesize intermediate video sequences conditioned on the given start and end frames.
Current state-of-the-art methods primarily extend large-scale pre-trained Image-to-Video Diffusion Models (I2V-DMs) by incorporating the end-frame condition via direct fine-tuning or temporally bidirectional sampling.
However, the former results in a weak end-frame constraint, while the latter inevitably disrupts the input representation of video frames, leading to suboptimal performance.
To improve the end-frame constraint while avoiding disruption of the input representation, we propose a novel video inbetweening framework specific to recent and more powerful transformer-based I2V-DMs, termed \textbf{EF-VI}.
It efficiently strengthens the end-frame constraint by utilizing an enhanced injection.
This is based on our proposed well-designed lightweight module, termed \textbf{EF-Net}, which encodes only the end frame and expands it into temporally adaptive frame-wise features injected into the I2V-DM.
Extensive experiments demonstrate the superiority of our EF-VI compared with other baselines.
\end{abstract}    
\section{Introduction}
% \begin{figure*}[ht]
%     \centering
%     \includegraphics[width=1.0\linewidth]{figure/figurea1.png}
%     \caption{\textbf{Comparison between current I2V-DM-based methods and our EF-VI.}
%     All the methods are based on the same I2V-DM~\cite{yang2025cogvideox}, CogVideoX-5B-I2V. 
% (a)~Ground truth.
% (b)~I2V Results (conditioned by the start frame only) of the base model.
% (c)~Video inbetweening results of directly-finetuning-based method.
% (b)~Video inbetweening results of temporally-bidirectional-based method.
% (e)~Video inbetweening results of our EF-VI.
% }
% \label{pipe_comparison}
% \end{figure*}

Video inbetweening aims to predict intermediate content conditioned on the given start and end frames, which is significant in film production~\cite{reda2022film}, animation creation~\cite{xing2024tooncrafter, zhu2025thin}, and video content editing~\cite{tanveer2024motionbridge}.
Traditional methods for video inbetweening (interpolation)~\cite{niklaus2020softmax, zhang2023extracting, liu2024sparse, zhang2025eden} mainly train a neural network from scratch to estimate intermediate optical flow, so they might only work for simple rigid motions with narrow gaps of input frames.

Since large pre-trained image-to-video diffusion models (I2V-DMs)~\cite{blattmann2023stable, xing2024dynamicrafter, yang2025cogvideox, wang2025wan} (especially transformer-based I2V-DMs) have general generative priors, recent works utilize them for video inbetweening~\cite{danier2024ldmvfi}, enabling larger dynamics and diverse predictions of intermediate content.
For the process of the end-frame condition, current I2V-DM-based methods have converged into two types.
The first type naively finetunes the I2V-DMs~\cite{wang2024framer, yang2025cogvideox} by injecting the start and end frames simultaneously in the same manner.
However, the original large-scale I2V-pretraining is only applicable to the start-frame condition, and the training specialized for the end-frame condition is significantly less sufficient.
This causes the control strength of the end frame over the generated intermediate content to be weaker than that of the start frame.
As shown in Fig.~\ref{pipe_comparison}~(a)-(c), the rider's motions in the intermediate frames generated by this method are closer to the image-to-video dynamics of the base model (conditioned on the start frame only) while deviating from the expected ground truth (conditioned on both the start and end frames), leading to degradation in the generated content.

To improve the end-frame constraint for better generation, another type combines a Unet-based pre-trained I2V-DM~\cite{blattmann2023stable} with a temporally bidirectional sampling strategy~\cite{wang2024generative, yang2025vibidsampler}, utilizing additional steps to denoise temporally reversed video only conditioned on the end frame.
However, this reverses the temporal order of frames, which may lead to dynamics violating real-world physical laws, such as ``upward-flowing water" or ``a person suddenly walking backward".
Furthermore, recent and more powerful transformer-based I2V-DMs utilize causal 3D-VAEs~\cite{yu2024language} for temporal compression of video frames.
Reversing these temporally compressed frames will further disrupt the input representation.
When implementing these methods into more recent base models, the generated content will suffer more severe corruption, as shown in Fig.~\ref{pipe_comparison}~(d).

Since enhancing the end-frame constraint means enabling the generation of intermediate frames to be strongly influenced by the end frame, this is similar to controllable image~\cite{peng2024controlnext} and video~\cite{zhang2023controlvideo, lin2024ctrl} generation at the target level: both enforce the output to be constrained by additional conditions.
The most efficient framework design for this task is to utilize an additional network to encode the pixel-wise or frame-wise control signals and explicitly inject them into the base models via direct addition or cross-attention~\cite{vaswani2017attention}.
Hence, borrowing their paradigms to inject the end-frame condition specifically would likely efficiently strengthen its constraint, bringing enhancement to our task.
However, different from then, we only have an end-frame condition instead of frame-wise control signals.
Temporally extending the end-frame condition to frame-wise features before injecting it into the base model should be further explored.

Based on this insight, we present \textbf{EF-VI}, a novel video inbetweening framework specific to the transformer-based I2V-DM.
Specifically, it processes the start frame in the same manner as the I2V-DM while enhancing the injection of the end-frame condition.
This is achieved by our well-designed lightweight module, \textbf{EF-Net}, which efficiently encodes only the end frame and expands it into temporally adaptive frame-wise features explicitly injected into the I2V-DM.
It enhances the end-frame constraint while avoiding disruption of the input representation, enabling our EF-VI to produce more harmonious transitions in the challenging scenario compared to other methods, as shown in Fig.~\ref{pipe_comparison}~(e).

Our main contributions are summarized as follows:\\
(1) We propose a novel video inbetweening framework specific to the transformer-based I2V-DM, termed \textbf{EF-VI}, which efficiently enhances the end-frame constraint through improved condition injection while avoiding disruption of the input representation.
(2) To achieve the enhanced injection, we propose a lightweight module, termed \textbf{EF-Net}, which efficiently encodes only the end frame and expands it into temporally adaptive frame-wise features injected into the I2V-DM.
(3) Extensive experiments demonstrate the superiority of our method compared with other baselines.

\section{Related Work}

\textbf{Traditional Video Inbetweening.}
Video inbetweening is a widely researched computer vision task that generates intermediate frames based on the start and end frames. 
Most traditional interpolation methods first train a network to estimate the optical flows between the start and end frames~\cite{huang2022real, reda2022film}.
Then, they synthesize the middle frames by utilizing the predicted optical flow to warp or splat~\cite{li2023amt, zhang2023extracting}.
However, due to the limitations of optical flow in temporal modeling~\cite{wu2024perception, liu2024video}, they fail to produce acceptable results when facing large motions or complex scenarios.
Besides, given an input pair, they can not provide multiple results due to deterministic predictions.

\textbf{Generative Video Inbetweening.}
With the development of large pre-trained image-to-video diffusion models (I2V-DMs)~\cite{blattmann2023stable, xing2024dynamicrafter} (especially recent and more powerful transformer-based I2V-DMs~\cite{yang2025cogvideox, wang2025wan}), recent works utilize their general generative priors~\cite{rombach2022high, peebles2023scalable, flux2024} for video inbetweening~\cite{danier2024ldmvfi, zhang2025eden} and achieve state-of-the-art results.
They leverage I2V-DMs by directly finetuning I2V-DMs without significant modification of the network structure, or utilizing a temporally bidirectional sampling strategy~\cite{zhu2024generative, wang2024generative, feng2024explorative}.
However, due to a large gap in the training scale between the start and end frames, the former has a weak end-frame constraint, while the latter inevitably disrupts the input representation of video frames, which may lead to dynamics violating real-world physical laws and distorted input representations of the base model.
 \begin{figure*}[ht]
     \centering
     \includegraphics[width=1.0\linewidth]{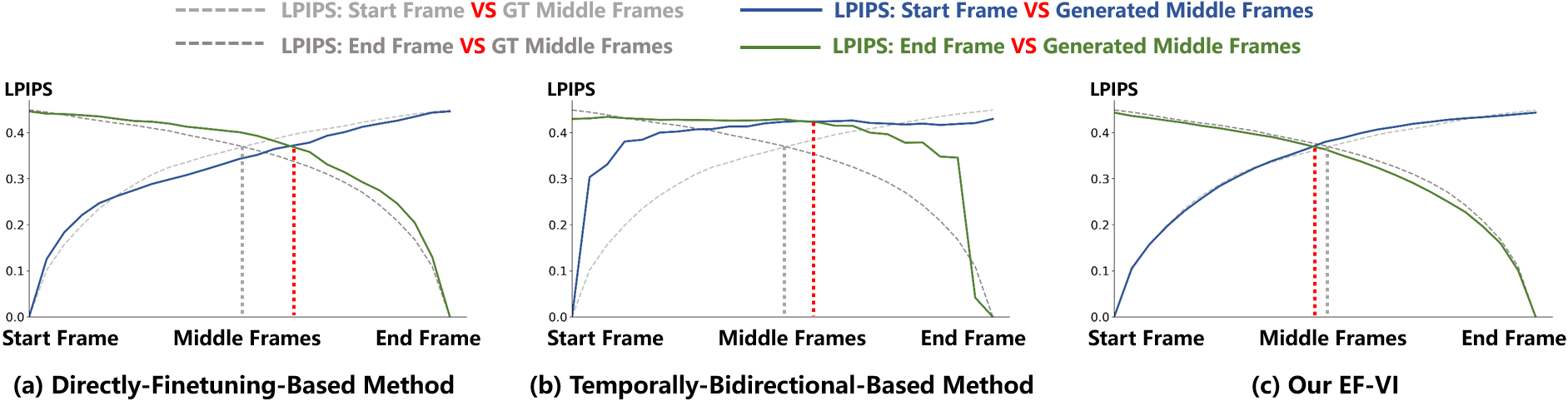}
     \caption{\textbf{LPIPS curves of current I2V-DM-based methods, our EF-VI, and the Ground Truth (GT).}
     For each method and the GT, we calculate the average LPIPS~\cite{zhang2018unreasonable} (a lower value indicating greater similarity) between the generated intermediate frames and the two boundary frames on 119 test pairs from the DAVIS dataset~\cite{pont20172017}.
 (a) Directly-Finetuning-Based Method and GT.
 (b) Temporally-Bidirectional-Based Method and GT.
 (c) Our EF-VI and GT.
}
\label{pipe_comparison_lpips}
\end{figure*}

\textbf{Controllable Image and Video Generation.}
To improve the controllability and stability of image and video generation, some methods~\cite{zhang2024tora, zhang2025easycontrol, he2025cameractrl} incorporate additional control signals, such as depth~\cite{esser2023structure}, trajectory~\cite{wang2024framer}, 3D point cloud~\cite{liu2024reconx}, match lines and human pose~\cite{zhu2024generative}, and optical flow~\cite{zhang2025motion}, into the base diffusion models.
The most efficient framework design for this task is to utilize an additional network to encode the pixel-wise or frame-wise control signals and explicitly inject them into the base models via direct addition or cross-attention.
For video inbetweening, frame-wise conditions are difficult to obtain.
Thus, most of them are derived from interpolation~\cite{tanveer2024motionbridge, zhu2024generative} between the conditions from the start and end frames or estimation from a traditional model~\cite{zhang2025motion}.
These interpolation-based frame-wise conditions probably conflict with real-world motions, degrading the generated content.
\section{Method}
\label{sec::3}
% In this section, we first review the background of I2V-DMs.
% Then, we introduce the motivation and proposal of our \textbf{EF-VI}, built on recent and more powerful transformer-based I2V-DMs.
% After that, we provide details about the network design of \textbf{EF-Net}, the core module in EF-VI.
\begin{figure*}[t]
    \centering
    \includegraphics[width=1.0\linewidth]{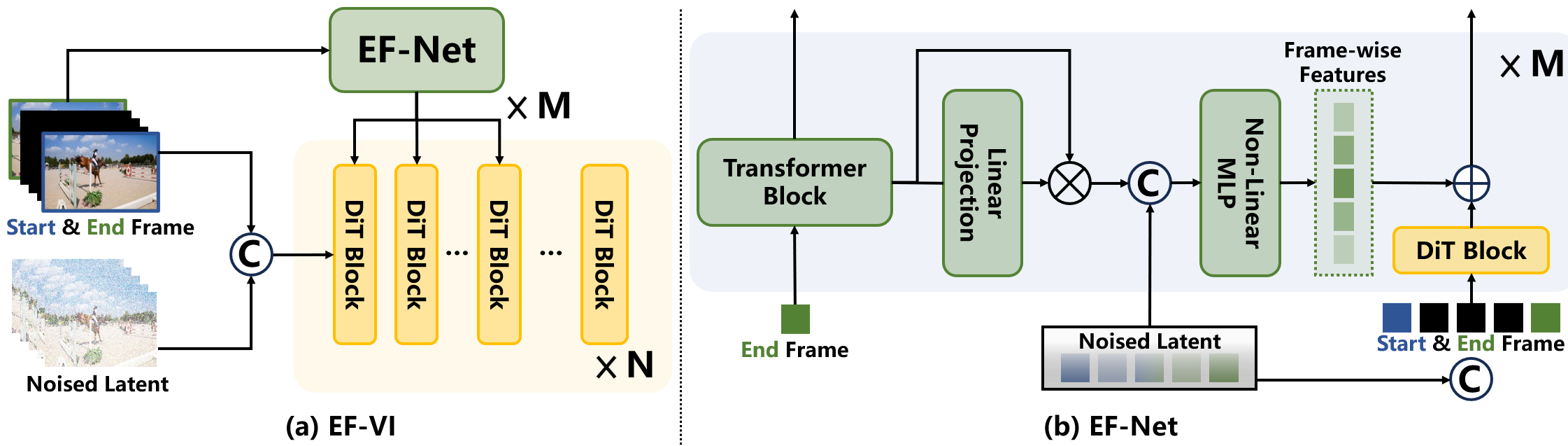}
    \caption{\textbf{Our proposed framework EF-VI and its core module EF-Net.}
(a) Our EF-VI efficiently enhances end-frame constraints by handling the start frame as before and applying an improved injection to the end frame.
(b) This new injection mechanism is based on a lightweight module, termed EF-Net, which efficiently encodes the end frame, expands it into temporally adaptive frame-wise features, and injects them into the I2V-DM.
}
\label{pipeline}
\end{figure*}

\subsection{Preliminaries: Image-to-Video Diffusion Model}
Image-to-Video Diffusion Model (I2V-DM) is mainly based on the Text-to-Video Diffusion Model (T2V-DM)~\cite{videoworldsimulators2024}, and enables video generation conditioned on the start frame.
When training, given an $F$-frame video $\boldsymbol{x} \in \mathbb{R}^{F\times 3\times H\times W}$, I2V-DM first uses a variational autoencoder~(VAE)~\cite{kingmaauto, van2017neural} to transform $\boldsymbol{x}$ into the latent representation $\boldsymbol{z} \in \mathbb{R}^{f\times c\times h\times w}$.
Then, random Gaussian noise $\boldsymbol{\epsilon}$ is added to $\boldsymbol{z}$, based on the noise schedule $\alpha_t$ and $\sigma_t$ at time $t$:
\begin{equation}
    \boldsymbol{z}_t = \alpha_t \boldsymbol{z} + \sigma_t \boldsymbol{\epsilon}, \quad \boldsymbol{\epsilon} \sim \mathcal{N}(\boldsymbol{0}, \mathbf{I}).
\end{equation}
The obtained $\boldsymbol{z}_t$ is used to train a denoiser $\mathcal{D}_\theta$ with parameter $\theta$, which denoises based on time $t$, image condition $\boldsymbol{c}_{i}$ (and optional other conditions $\boldsymbol{c}_o$) by minimizing the loss:
\begin{equation}
    \mathcal{L}=\mathbb{E}_{\boldsymbol{z}_t, t, \boldsymbol{\epsilon}, \boldsymbol{c}_{i}, \boldsymbol{c}_{o}} \left[ \left\lVert \mathcal{D}_\theta(\boldsymbol{z}_t; t, \boldsymbol{c}_{i}, \boldsymbol{c}_o) - \boldsymbol{y}_t \right\rVert_2^2 \right],
\end{equation}
where $\boldsymbol{y}_t$ is the target for the output of the network at time $t$, determined by a specific network prediction, such as original noise prediction~\cite{ho2020denoising} or $v$-prediction~\cite{salimansprogressive}.

For inference, we use the denoiser $\mathcal{D}_\theta$ to recover the latent representation from pure Gaussian noise $\boldsymbol{Z}_T \sim \mathcal{N}(\boldsymbol{0}, \mathbf{I})$ by iteratively denoising $T$ times.
Specifically, at time $t$, $\mathcal{D}_\theta$ will provide a prediction $\boldsymbol{\hat{y}}_t$ based on $\boldsymbol{z}_t$ (obtained from the previous denoising step), $t$, $\boldsymbol{c}_i$, and $\boldsymbol{c}_o$.
Then, it produces a cleaner noised latent $\boldsymbol{z}_{t-1}$ with less noise, through specific mathematical calculation $\mathcal{C}$~\cite{yang2025cogvideox}, expressed as:
\begin{equation}
    \boldsymbol{z}_{t-1} = \mathcal{C}(\boldsymbol{z}_t, \boldsymbol{\hat{y}}_t = \mathcal{D}_\theta(\boldsymbol{z}_t; t, \boldsymbol{c}_{i}, \boldsymbol{c}_o), t).
\end{equation}
After $T$ denoising steps, we use the same VAE to decode $\boldsymbol{z}_0$ and obtain the generated video.

\subsection{Motivation and Proposal of EF-VI}
From the perspective of processing the end-frame condition, current I2V-DM-based methods can be divided into two types.
The first type finetunes the I2V-DMs by using the same manner to inject the start and end frames ($\boldsymbol{c}_s$ and $\boldsymbol{c}_e$) simultaneously, such as channel concatenation with $\boldsymbol{z}_t$ or cross-attention with inner features (denoted as $\mathcal{J}(\cdot)$).
For simplification (ignoring unimportant conditions and calculations), this can be expressed as:
\begin{equation}
\label{eq4}
\boldsymbol{z}_{t-1} = \mathcal{D}_\theta(\boldsymbol{z}_t;t, \mathcal{J}(\boldsymbol{c}_s), \mathcal{J}(\boldsymbol{c}_e)).
\end{equation}
However, the original large-scale I2V-pretraining of their basic I2V-DMs is only for the start-frame constraint, which focuses on the following equation:
\begin{equation}
\label{eq5}
\boldsymbol{z}_{t-1} = \mathcal{D}_\theta(\boldsymbol{z}_t; t, \mathcal{J}(\boldsymbol{c}_s)).
\end{equation}
Naively incorporating the end-frame condition through the same injection manner $\mathcal{J}(\cdot)$ is far from making up for the gap in training scales, which causes the end-frame constraint to be weaker than the start-frame constraint.
We quantify this by calculating the average LPIPS~\cite{zhang2018unreasonable} (a lower value indicating greater similarity) between the generated middle frames and the two boundary frames on 119 test pairs from the DAVIS dataset~\cite{pont20172017}. 
As shown in Fig.~\ref{pipe_comparison_lpips}~(a), the LPIPS between the start frame and generated middle frames is generally lower than that of the ground truth (GT) middle frames, while the LPIPS between the end frame and generated middle frames is higher than that of the GT middle frames.
We consider that, compared to the ground truth, the generated video frames are more similar to the start frame and less like the end frame, reflecting a weaker end-frame constraint. 

To improve the end-frame constraint, another type combines a Unet-based pre-trained I2V-DM, SVD~\cite{blattmann2023stable}, with the temporally bidirectional sampling strategy.
The core of these methods is utilizing additional steps to denoise the temporally reversed video conditioned only on the end frame, which can be expressed as:
\begin{equation}
\label{eq6}
    \boldsymbol{z}_{t-1}^{s} = \mathcal{D}_\theta(\boldsymbol{z}_t; t, \mathcal{J}(\boldsymbol{c}_s)),
\end{equation}
\begin{equation}
\label{eq7}
    \boldsymbol{z}_{t-1}^{e} = \mathcal{D}_\theta(\textit{Flip}(\boldsymbol{z}_t); t, \mathcal{J}(\boldsymbol{c}_e)),
\end{equation}
\begin{equation}
\label{eq8}
    \boldsymbol{z}_{t-1} = \textit{Fuse}(\boldsymbol{z}_{t-1}^{s}, \boldsymbol{z}_{t-1}^{e}),
\end{equation}
where $\textit{Flip}(\cdot)$ denotes temporal reversal and $\textit{Fuse}(\cdot)$ is a fusion operation such as linear interpolation.
Although Eqs.~(\ref{eq6}-\ref{eq8}) theoretically make the end-frame constraint as strong as the start-frame constraint, the $\textit{Flip}(\cdot)$ reverses the temporal order of frames, which may lead to dynamics violating real-world physical laws.
Furthermore, recent and more powerful transformer-based I2V-DMs utilize causal 3D-VAEs~\cite{li2024wf} for temporal compression of video frames.
In this case, $\boldsymbol{z}_t$ has a strong temporal-forward causal relationship, and $\textit{Flip}(\boldsymbol{z}_t)$ falls outside the representation space of 3D-VAEs, detailed in Appendix A.
As shown in Fig.~\ref{pipe_comparison}~(b), the LPIPS values between the two boundary frames and generated middle frames are generally both much larger than those of the GT middle frames, indicating severe quality degradation of the generated content.

According to the above analysis, it is urgent to design a novel method for video inbetweening that enhances the end-frame constraint while avoiding the harmful $\textit{Flip}(\cdot)$ (especially for recent and more powerful transformer-based I2V-DMs).
Since controllable generation has been achieved by utilizing an additional network to encode and then inject the control signals into the base models, using an enhanced injection for the end-frame condition probably strengthens its influence on the intermediate content.
Based on this, we propose a novel video inbetweening framework, termed \textbf{EF-VI}, shown in Fig.~\ref{pipeline}~(a), built on a transformer-based I2V-DM, CogVideoX-5B-I2V.
It first processes the start frame $\boldsymbol{c}_s$ as in the base model by temporally zero-padding $\boldsymbol{c}_s$ and concatenating it with the noised latent $\boldsymbol{z}_t$ in the channel dimension.
Then, for the end-frame condition, besides the above processing as the start frame, we utilize a well-designed lightweight module, termed \textbf{EF-Net}, to enhance its injection into the base model.
The whole denoising process of the framework can be expressed as:
\begin{equation}
\label{eq9}
\boldsymbol{z}_{t-1} = \mathcal{D}_\theta(\boldsymbol{z}_t; t, \mathcal{J}(\boldsymbol{c}_s), \mathcal{J}(\boldsymbol{c}_e), \texttt{EF-Net}(\boldsymbol{c}_e))
\end{equation}
As shown in Fig.~\ref{pipe_comparison_lpips}~(c), compared to the other two types of methods, the LPIPS curves of our EF-VI are closer to those of the GT.
This proves that our method can achieve a more symmetric constraint of the start and end frames.
\begin{table*}[ht]
   \centering
   \scalebox{1.0}{
   \begin{tabular}{c|c|cccc|cccc}
     \hline
     \multirow{2}{*}{Methods}& \multirow{2}{*}{Base Model} &\multicolumn{4}{c}{DAVIS} &\multicolumn{4}{|c}{Pexels}  \\
     \cline{3-10}
         &&LPIPS$\downarrow$ &FID$\downarrow$ &FVD$\downarrow$ &VBench$\uparrow$ &LPIPS$\downarrow$ &FID$\downarrow$ &FVD$\downarrow$ &VBench$\uparrow$ \\
     \hline
     FILM &/&\underline{0.2313} &37.47 &739.07 &\underline{0.8162} &\underline{0.2254} &36.37 &734.43 &\underline{0.8261}  \\
     EMA-VFI &/&0.3697 &74.02 &882.44 &0.7342 &0.3270 &63.08 &804.39 &0.7671 \\
     \hline
     Framer-FT &SVD&0.2586 &45.65 &502.33 &0.7573 &0.2532 &39.86 &634.93 &0.7945\\
     GI-BD &SVD&0.2463 &31.64 &555.60 &0.7929 &0.2511 &32.51 &650.59 &0.8140\\
     FCVG-BD &SVD&0.2467 &29.24 &531.00 &0.8000 &0.2440 &30.04 &607.25 &0.8166 \\
     \hline
     CogVX-FT &CogVideoX-5B-I2V&0.2349 &\underline{26.46} &\underline{449.02} &0.8104 &0.2376 &\underline{27.96} &\underline{526.42} &0.8214\\
     CogVX-BD &CogVideoX-5B-I2V&0.3864&71.01&937.26&0.7768&0.3381&66.24&736.45&0.7931\\
     %Wan2.1-FLF2V & & & & & & & &   \\
     %\midrule
     \textbf{Ours} &CogVideoX-5B-I2V&\textbf{0.2096} &\textbf{22.30} &\textbf{382.03} &\textbf{0.8240} &\textbf{0.2246} &\textbf{24.50} &\textbf{501.70} &\textbf{0.8373}  \\
     \hline
   \end{tabular}
   }
   \caption{\textbf{Quantitative comparison between our EF-VI and some state-of-the-art methods on DAVIS and Pexels datasets.}
   The best and second-best scores for each metric are \textbf{bolded} and \underline{underlined}, respectively.}
   \label{quantitative comparison}
 \end{table*}

\subsection{Network Design of EF-Net}
To enhance the injection of the end-frame condition, we propose our EF-Net, inspired by successful network design paradigms in controllable image and video generation, such as ControlNet~\cite{zhang2023adding} and T2I-Adapter~\cite{mou2024t2i}.
However, these previous modules for adding control to the base model can only handle pixel-wise or frame-wise conditions.
In contrast, we convert a single image (the end frame) into temporally adaptive frame-wise features before injecting them into the I2V-DM.
The network structure of EF-Net is shown in Fig.~\ref{pipeline}~(b).
Specifically, it takes only the end frame $\boldsymbol{c}_e$ as input and transforms it into $M$ serial features $\boldsymbol{F}_j\in \mathbb{R}^{L\times D}~(1\leq j\leq M)$ by a series of transformer blocks $\mathcal{B}_j~(1\leq j\leq M )$, expressed as:
\begin{equation}
\label{eq10}
    \boldsymbol{F}_j=\mathcal{B}_j(\boldsymbol{F}_{j-1}),
\end{equation}
where $\boldsymbol{F}_0$ is obtained by patchifying $\boldsymbol{c}_e$ into tokens.
The $L$ and $D$ are the number of tokens and the token dimension, respectively.
To turn $\boldsymbol{F}_j$ into frame-wise features, we first utilize a linear projection $\mathcal{P}_j$ that predicts token-wise temporal coefficients.
These coefficients can temporally expand $\boldsymbol{F}_j$ by $f$ times (the frame number of $\boldsymbol{z}_t$) by an outer product operation, which can be expressed as:
\begin{equation}
\label{eq11}
    \hat{\boldsymbol{F}}_j=\mathcal{P}_j(\boldsymbol{F}_j) \times \boldsymbol{F}_j,
\end{equation}
where $\mathcal{P}_j(\boldsymbol{F}_j) \in \mathbb{R}^{L\times f}$ and $ \hat{\boldsymbol{F}}_j \in \mathbb{R}^{(L\times f)\times D}$.
To make $\hat{\boldsymbol{F}}_j$ more temporally adaptive, we should incorporate dynamic information of the entire video sequence into this end-frame feature.
Actually, the temporal information already explicitly exists in the latent $\boldsymbol{z}_t$, which represents the video content, though interfered with by noise.
Hence, to incorporate $\boldsymbol{z}_t$, we patchify then concatenate it with $\hat{\boldsymbol{F}}_j$.
This composite is fed into a non-linear MLP to produce the final frame-wise feature $\overline{\boldsymbol{F}}_j$, expressed as:
\begin{equation}
\label{eq12}
    \overline{\boldsymbol{F}}_j=\texttt{MLP}(\textit{Concat}(\hat{\boldsymbol{F}}_j,\textit{Patchify}(\boldsymbol{z}_t))).
\end{equation}
After that, we directly add $\overline{\boldsymbol{F}}_j$ to the features output by the first $M$ blocks of the I2V-DM:
\begin{equation}
\label{eq13}
\boldsymbol{A}_{j+1}=\mathcal{B}^{\text{I2V}}_j(\boldsymbol{A}_{j})+\overline{\boldsymbol{F}}_{j},
\end{equation}
where $\mathcal{B}^{\text{I2V}}_j$ denotes the $j$-th block of the I2V-DM and $\boldsymbol{A}_{j}$ is its input.
To ensure the lightness of EF-Net, we set $M=4$, which is much smaller than the number of blocks in the I2V-DM ($N=42$).
Transformer blocks $\mathcal{B}_{j}~(1\leq j\leq M )$ in our EF-Net can vary with different network structures.
In this work, we utilize the DiT block in CogVideoX-5B-I2V for convenience.
We consider this sufficient to impose a strong end-frame constraint, since the end-frame features have explicitly influenced the inner features of the entire video in the base I2V-DM by frame-wise condition injection.
\section{Experiments}
\label{sec::4}
%In this section, we first introduce our experimental settings.
%Then, we compare our EF-VI with other state-of-the-art methods, and further ablation experiments for our method are presented.
%Finally, we show the generalization ability of our EF-VI to the cartoon video inbetweening.

 \begin{figure*}[ht]
    \centering
    \includegraphics[width=1.0\linewidth]{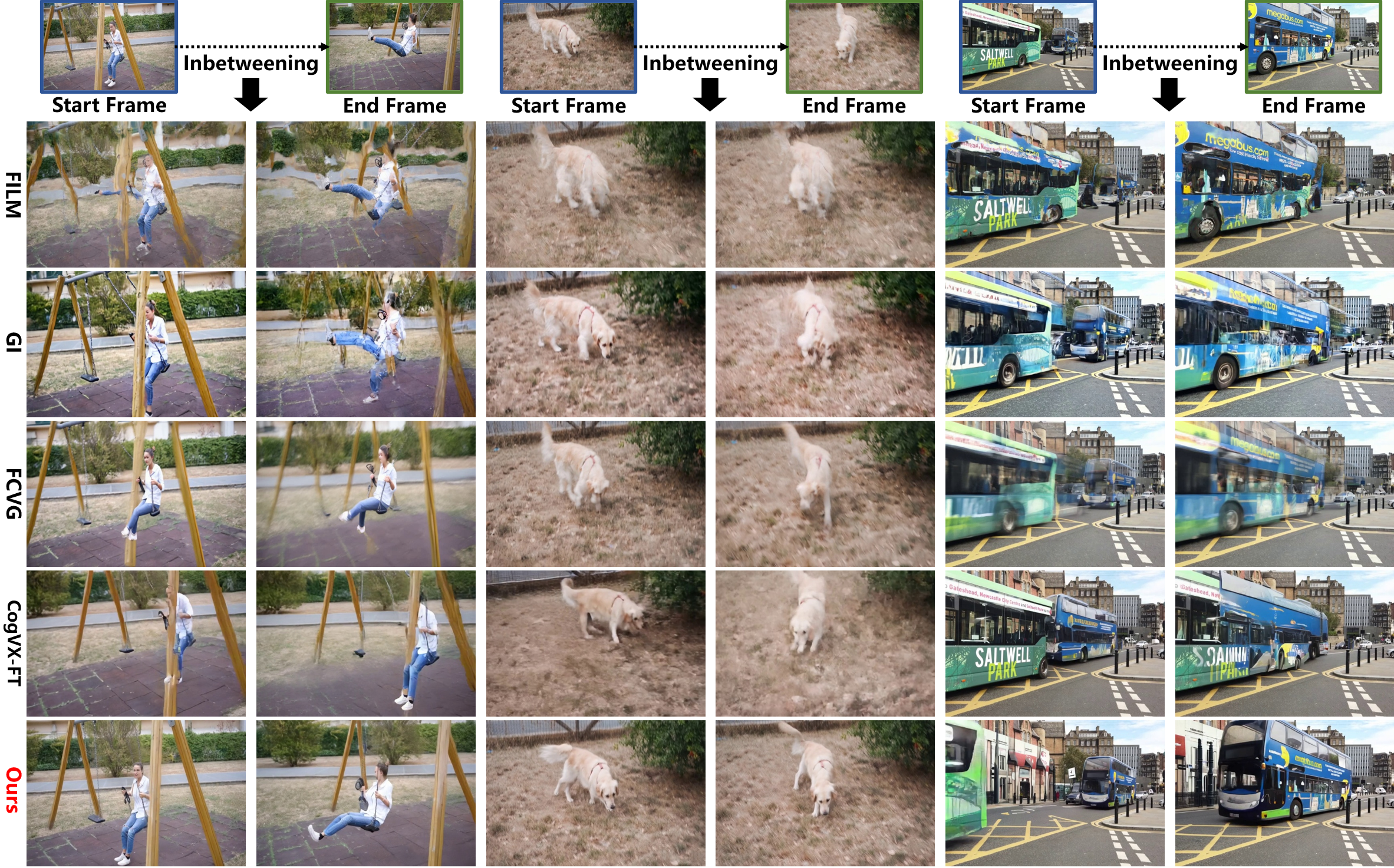}
    \caption{\textbf{Visual comparison between our EF-VI and some state-of-the-art methods.}
    Our EF-VI significantly improves the visual quality of the generated intermediate content across diverse scenarios.
}
\label{visual_comparison}
\end{figure*}
\begin{figure*}[t]
    \centering
    \includegraphics[width=1.0\linewidth]{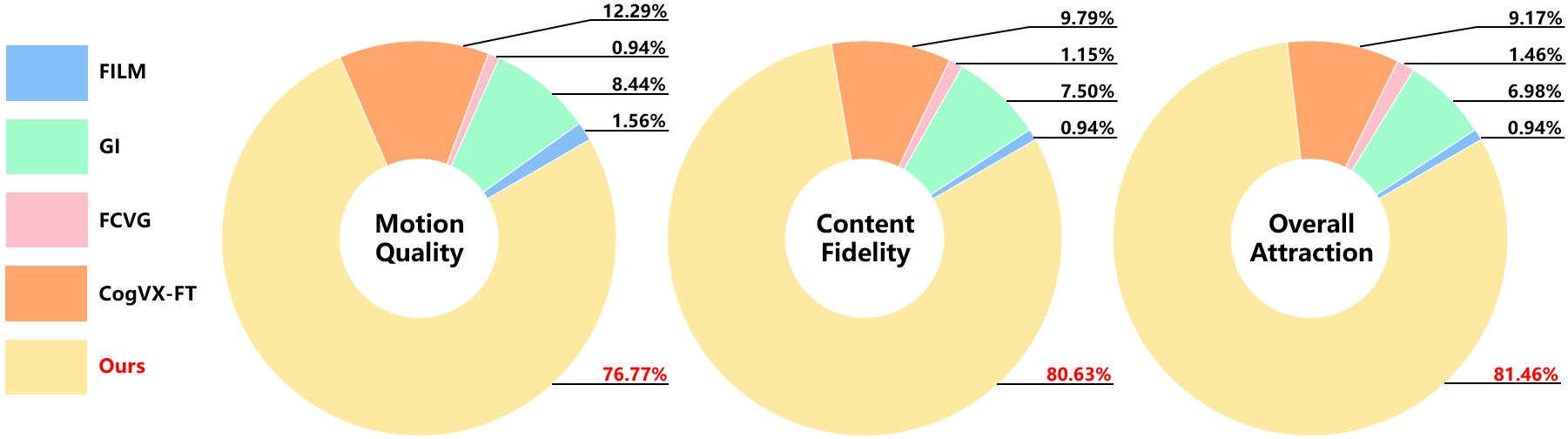}
    \caption{\textbf{Results of the user study.}
    Each pie chart illustrates the proportion of videos output by each method that are selected by participants in a specific evaluation dimension.
}
\label{user_study}
\end{figure*}

\subsection{Experimental Setting}
\label{sec::4.1}
\textbf{Datasets.}
We obtain the training data by collecting real-world videos from the publicly available creative material platform iStock.
For testing, we follow most related works on generative video inbetweening~\cite{wang2024generative, yang2025vibidsampler}.
Specifically, we curate 119 and 100 video clips from the DAVIS dataset~\cite{pont20172017} and Pexels, respectively.
The video clips in the two test datasets cover various scenarios, including human actions, animal motions, vehicle movements, and natural scenes.

\textbf{Implementation Details.}
We train our EF-VI model with low resource consumption.
Specifically, the training process is conducted on 4 NVIDIA A800 GPUs, requiring only $6k$ iterations with a total batch size of 4.
%Specifically, the training process requires only $6k$ iterations, with a total batch size of 4.
The AdamW optimizer~\cite{kingma2014adam} is used to update the parameters of both EF-Net and the entire base model (CogVideoX-5B-I2V) simultaneously.
We employ a cosine annealing learning rate with an initial value of 3e-5.
The number of inference steps for EF-VI is 50, matching the officially recommended setting for CogVideoX-5B-I2V.
%The number of inference steps for EF-VI is 50, matching the officially recommended setting.
%(Xiaoyu: can not mention controlled GPU on open-published version)

\textbf{Evaluation Metrics.}
Following most related works, we adopt LPIPS~\cite{zhang2018unreasonable} and FID~\cite{heusel2017gans} to assess the individual frame quality.
Additionally, we utilize FVD~\cite{unterthiner2019fvd} and a recently proposed benchmark, VBench~\cite{huang2024vbench}, to measure the overall quality of the videos.
Since VBench evaluates videos across multiple dimensions, following~\cite{wang2024generative, zhang2025motion}, we use the officially provided weights to calculate the final weighted average scores.
More details of evaluation metrics can be found in Appendix B.

\subsection{Comparison with Other Baselines}
\textbf{Baselines.}
We quantitatively compare our EF-VI with 7 advanced baselines for video inbetweening.
Specifically, FILM~\cite{reda2022film} and EMA-VFI~\cite{zhang2023extracting} are widely used optical-flow-based methods.
Framer is finetuned from an Unet-based I2V-DM, SVD~\cite{blattmann2023stable}, with additional trajectory control.
GI~\cite{wang2024generative} and FCVG~\cite{zhu2024generative} are also built on SVD but adopt the temporally bidirectional sampling strategy.
CogVX-FT~\cite{CogVideoX-FT} is an open-source project for video inbetweening that directly finetunes CogVideoX-5B-I2V~\cite{yang2025cogvideox} \textbf{with 10 times more training overhead than ours}.
We also self-implement the temporally bidirectional sampling in CogVideoX-5B-I2V for video inbetweening, denoted as CogVX-BD.
For further qualitative comparison and user study, we select relatively well-performing methods from each category based on their quantitative results, including FILM (optical-flow-based), GI and FCVG (temporally-bidirectional-based), as well as CogVX-FT (directly-finetuning-based).

\textbf{Quantitative Comparison.}
Since the generation processes of methods built on CogVideoX-5B-I2V are controlled not only by the start and end frames but also by the text prompts, for a fair comparison, we utilize a popular vision language model, Qwen2.5-VL-7B~\cite{Qwen2.5-VL}, to automatically predict the text prompts only based on the start and end frames, detailed in Appendix C.
For a clearer representation, we add "FT (directly-finetuning-based)" or "BD (temporally-bidirectional-based)" to SVD-based methods as a suffix.
As shown in Tab.~\ref{quantitative comparison}, our EF-VI achieves the highest scores across all metrics on the DAVIS and Pexels datasets.
On both datasets, the LPIPS, FID, FVD, and VBench of our EF-VI are much better than those of the CogVX-FT and CogVX-BD, which utilize the same base I2V-DM as ours.
Besides, when SVD serves as the base model, compared to direct finetuning, temporally-bidirectional-based methods can obtain better results.
However, the results are opposite for methods built on the transformer-based I2V-DM, since its input has a strong temporal-forward causal relationship, and the temporally bidirectional sampling will disrupt the input representation.
Although these metrics do not fully align with human intuitive perception, they still prove the significance of our enhanced end-frame injection for I2V-DM-based video inbetweening methods, to some degree.

\textbf{Qualitative Comparison.}
The superiority of our method is further reflected in Fig.~\ref{visual_comparison}.
When the start and end frames have a large gap, other methods produce inharmonious intermediate content with inconsistent motions or collapsed appearances.
However, our EF-VI can still provide smooth transitions.
For example, as shown in the first two columns of Fig.~\ref{visual_comparison}, the intermediate frames generated by other methods all contain inappropriate movement trajectories of the person or distorted content.
On the contrary, our EF-VI achieves a much better result with harmonious dynamics.
More visual results of our EF-VI are shown in Appendix D.

\textbf{User Study.}
We conduct a user study to measure human preference for model outputs across three dimensions: \textbf{motion quality}, \textbf{content fidelity}, and \textbf{overall attraction}.
%Motion quality means whether the motions in the video are smooth and conform to the real-world physical laws.
%Content fidelity means whether the contents in the video frames are consistent and undistorted.
%Overall attraction means whether the video is attractive to people as a whole.
Specifically, we utilize EF-VI and four other methods to produce corresponding results for 30 start-end frame pairs.
Then, for videos with the same start and end frames, participants separately selected the best ones based on the three dimensions.
A total of 32 participants took part in this experiment, providing 2,880 ratings.
As illustrated in Fig.~\ref{user_study}, for each evaluated dimension, the proportion of our EF-VI being selected exceeds three-quarters, demonstrating a strong human preference for the outputs produced by our method.
More details of the user study are provided in Appendix E.

\begin{table}[t]
   \centering
   \scalebox{1.0}{
   \begin{tabular}{cccc}
     \hline
      Methods &Video Size &Step &Time~(s) \\
      \hline
      Framer &25$\times$1024$\times$576&30&61.56 \\
     GI &25$\times$1024$\times$576&50&616.52 \\
     FCVG &25$\times$1024$\times$576&25&115.44 \\
     CogVX-FT &49$\times$720$\times$ 480&50&234.89 \\
     CogVX-BD &49$\times$720$\times$ 480&50&475.28 \\
     \textbf{Ours} &49$\times$720$\times$ 480&50&237.82 \\
     \hline
   \end{tabular}
   }
   \caption{\textbf{Inference times of I2V-DM based methods on a single NVIDIA A800 80G GPU.}
    Our EF-VI utilizes the same base I2V-DM as CogVX-FT with enhanced end-frame injection but only increases a little inference time.}
    \label{efficiency}
 \end{table}

\textbf{Inference Efficiency.}
Besides, we test the inference times of our EF-VI and other I2V-DM-based methods on a single NVIDIA A800 80G GPU.
%Besides, we test the inference times of our EF-VI and other I2V-DM-based methods.
We keep the denoising steps for other methods the same as their official implementations.
As shown in Tab.~\ref{efficiency}, methods based on different I2V-DMs will produce videos of different sizes.
Although our EF-VI utilizes the same base I2V-DM as CogVX-FT, it only increases very little inference time, indicating the efficiency of our proposed method.
In contrast, CogVX-BD doubles the inference time for additional temporally backward sampling.

\begin{table}[t]
        \centering
        \scalebox{0.97}{
        \begin{tabular}{ccccc}
            \hline
             Variants &LPIPS$\downarrow$ &FID$\downarrow$ &FVD$\downarrow$ &VBench$\uparrow$  \\
            \hline
            w/o EF-Net &0.2270 &25.18 &422.35 &0.8096 \\
            EF-Net~(w/o $\boldsymbol{z}_t$) &0.2188 &22.81 &400.37 &0.8168\\
            EF-Net~(w/ $\boldsymbol{E}_j$) &0.2148 &24.89 &\textbf{375.89} &0.8151 \\
            \textbf{Ours} &\textbf{0.2096} &\textbf{22.30} &382.03 &\textbf{0.8240} \\
            \hline
        \end{tabular}}
        \caption{\textbf{Different module designs of EF-Net.}}
    \label{EF_Net_ablation}
\end{table}

\begin{table}[t]
        \centering
        \scalebox{1.0}{
        \begin{tabular}{ccccc}
            \hline
            Strength &LPIPS$\downarrow$ &FID$\downarrow$ &FVD$\downarrow$ &VBench$\uparrow$ \\
            \hline
            $w=0.5$ &0.2167 &23.72 &399.69 &0.8206 \\
            $w=1.0$ &\textbf{0.2096} &\textbf{22.30} &\textbf{382.03} &0.8240 \\
            $w=1.5$ &0.2147 &23.69 &430.21 &0.8243 \\
            $w=2.0$ &0.2204 &24.36 &447.31 &\textbf{0.8248} \\
            \hline
    \end{tabular}
    }
    \caption{\textbf{Scaling the frame-wise features of EF-Net.}}
    \label{Scaling}
\end{table}

\subsection{Ablation Study}
\textbf{Module Design of EF-Net.}
We train three other model variants with the same settings as EF-VI and compare these variants with EF-VI on the DAVIS dataset:
(1) \textbf{w/o EF-Net}: we directly remove the entire EF-Net.
(2) \textbf{EF-Net~(w/o $\boldsymbol{z}_t$}): we remove the incorporation of the noised latent $\boldsymbol{z}_t$~(Eq.~\ref{eq12}) in our EF-Net.
(3) \textbf{EF-Net~(w/ $\boldsymbol{E}_j$}): we consider adding a learnable temporal position embedding to temporally vary frame-wise features after the outer product operation in Eq.~\ref{eq11}.
According to Tab.~\ref{EF_Net_ablation}, removing the entire EF-Net leads to the poorest results, indicating the significance of using the additional module to enhance end-frame injection. 
Compared with EF-Net~(w/o $\boldsymbol{z}_t$) and EF-Net~(w/ $\boldsymbol{E}_j$), our method obtains the best results on LPIPS, FID, and VBench, except for a slightly lower FVD than that of the EF-Net~(w/ $\boldsymbol{E}_j$).
This indicates the effectiveness of incorporating the noised latent while omitting the temporal position embedding in our EF-Net's design.

\textbf{Scaling Frame-wise Features.}
For some modules used in controllable generation, such as ControlNet~\cite{zhang2023adding} and ControlNeXt~\cite{peng2024controlnext}, constraint strength can be manually adjusted by utilizing a factor $w$ to scale the pixel-wise or frame-wise features before adding them to the base model.
We also explore this scaling for the frame-wise features produced by our EF-Net.
According to the results on the DAVIS dataset shown in Tab.~\ref{Scaling}, naive scaling may degrade the performance of our EF-VI.
Thus, maintaining $w=1.0$ (aligned with the training setting) is probably the first choice in most scenarios.

\subsection{Generalization to Cartoon Video Inbetweening}
Video inbetweening is also a significant technology for cartoon creation ~\cite{siyao2023deep, shen2024bridging, zhong2024clearer}.
Although our EF-VI is trained on real-world data, it demonstrates strong generalization ability to cartoon video inbetweening.
As shown in Appendix F, our EF-VI can still obtain visually high-quality results even for challenging scenarios, such as complex actions of cartoon characters or animals, and large movements of vehicles or cameras.
More comparisons between our EF-VI and other baselines for cartoon video inbetweening are also provided in Appendix F.
\section{Limitations}
Our method's performance is limited by the generation capability of its base model (CogVideoX-5B-I2V).
Maintaining consistent dynamics and appearances remains challenging in scenarios involving fast or large-scale human movements and the motion of small objects.
A potential approach for improvement is to scale up the model size at the cost of much more computation, exemplified by the recently proposed industrial model Wan2.1-FLF2V-14B~\cite{wang2025wan}, which was developed concurrently with our work.
Comparisons between our method and Wan2.1-FLF2V-14B are also presented in Appendix G.

\section{Conclusion}
In this paper, to improve the end-frame constraint while avoiding disruption of the input representation, we proposed a novel video inbetweening framework specific to recent and more powerful transformer-based I2V-DMs, termed \textbf{EF-VI}.
It efficiently strengthened the end-frame constraint by utilizing an enhanced injection.
We developed this new injection by proposing a well-designed lightweight module, named~\textbf{EF-Net}, which could efficiently encode only the end frame and expand it into temporally adaptive frame-wise features injected into the I2V-DM.
Comprehensive experiments on video inbetweening demonstrated the superiority of our method compared to other baselines.
{
    \small
    \bibliographystyle{ieeenat_fullname}
    \bibliography{main}
}
\clearpage
\setcounter{page}{1}
\maketitlesupplementary

  \begin{figure*}[t]
    \centering
    \includegraphics[width=1.0\linewidth]{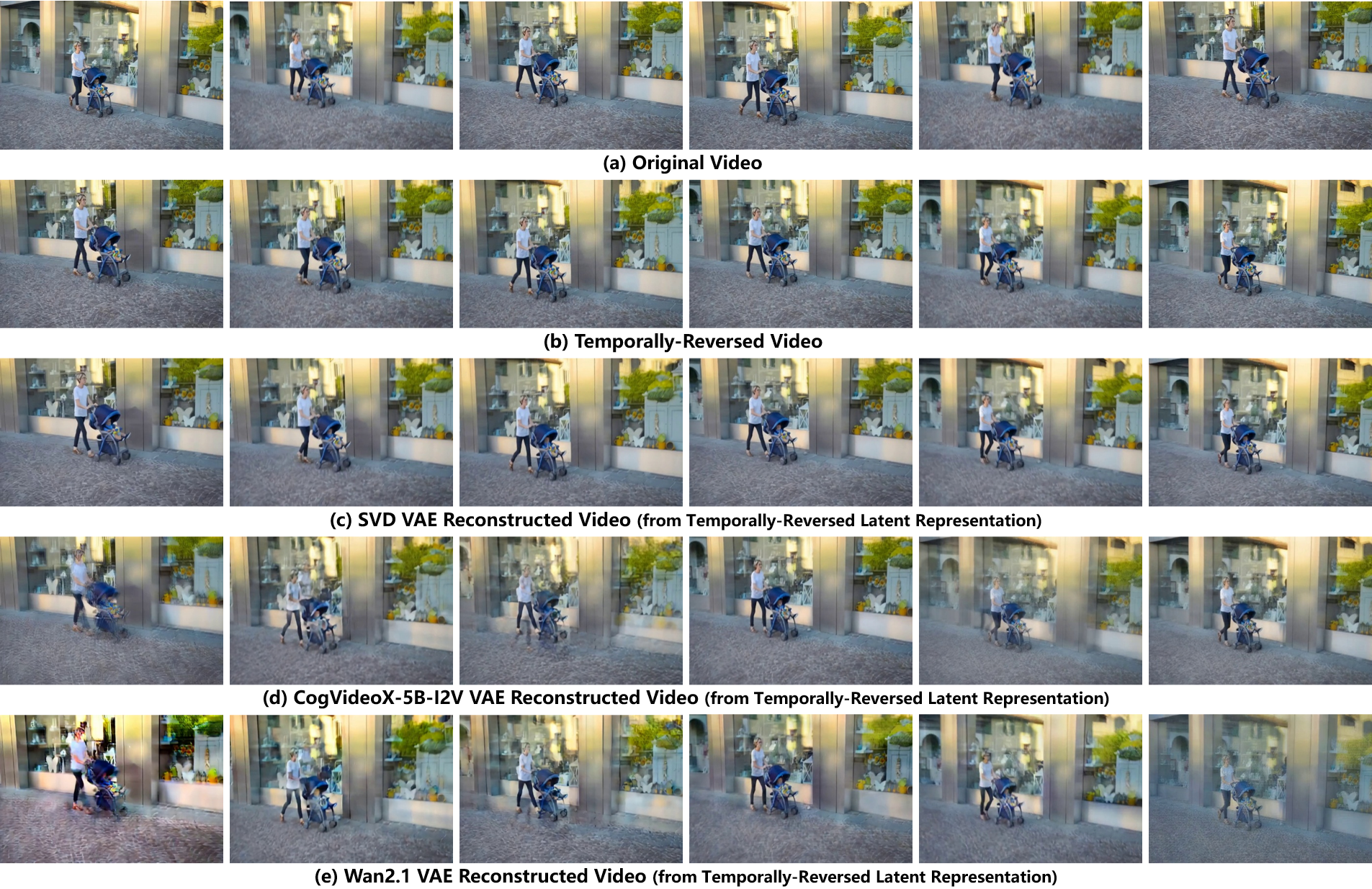}
    \caption{\textbf{Comparison of using different VAEs to reconstruct videos from the temporally-reversed latent representation.}
 (a)~Original video.
 (b)~Temporally-reversed video.
 (c)~Video reconstructed using SVD VAE from the temporally-reversed latent representation.
 (d)~Video reconstructed using CogVideoX-5B-I2V VAE from the temporally-reversed latent representation.
 (e)~Video reconstructed using Wan2.1 VAE from the temporally-reversed latent representation.}
\label{vae}
\end{figure*}
%\input{../../ReproducibilityChecklist/LaTeX/ReproducibilityChecklist.tex}
\begin{table*}[t]
   
   \centering
   \scalebox{1.0}{
   \begin{tabular}{cccccccc}
     \hline
     \multirow{2}{*}{Methods} &Subject &Background &Aesthetic &Imaging &Motion &Temporal &Average
     \\
       &Consistency &Consistency &Quality &Quality &Smoothness &Flickering &Score\\
     \hline
     FILM &0.9301 &0.9294 &0.5147 &0.6732 &0.9896 &0.9679 &0.8162\\
     EMA-VFI &0.8496 &0.8892 &0.4160 &0.3918 &\textbf{0.9906} &\textbf{0.9802} &0.7342\\
     Framer &0.8857 &0.9049 &0.4691 &0.6096 &0.9782 &0.9533 &0.7707 \\
     GI &0.9334 &0.9240 &0.5157 &0.6560 &0.9807 &0.9347 &0.7929\\
     FCVG &0.9269 &0.9262 &0.5139 &0.6405 &0.9829 &0.9558 &0.8000\\
     CogVX-FT &0.9280 &0.9387 &0.5327 &0.6692 &0.9834 &0.9539 &0.8104\\
     CogVX-BD &0.8447 &0.9084 &0.4909 &0.6327 &0.9767 &0.9681 &0.7768\\
     \hline
     \textbf{Ours} &\textbf{0.9388} &\textbf{0.9450} &\textbf{0.5416} &\textbf{0.6757} &0.9897 &0.9627  &\textbf{0.8240}\\
     \hline
   \end{tabular}
   }
   \caption{\textbf{VBench dimensional scores of our EF-VI and other baselines on the DAVIS dataset.}}
   \label{Vbench on DAVIS}
 \end{table*}
 \begin{table*}[t]
   \centering
   \scalebox{1.0}{
   \begin{tabular}{cccccccc}
     \hline
     \multirow{2}{*}{Methods} &Subject &Background &Aesthetic &Imaging &Motion &Temporal &Average
     \\
       &Consistency &Consistency &Quality &Quality &Smoothness &Flickering &Score\\
     \hline
     FILM &0.9356 &0.9380 &0.5291 &0.6855 &0.9906 &0.9720 &0.8261\\
     EMA-VFI &0.8791 &0.9155 &0.4626 &0.4593 &\textbf{0.9921} &\textbf{0.9831} &0.7671\\
     Framer &0.9039 &0.9248 &0.4995 &0.6444 &0.9802 &0.9616 &0.7945 \\
     GI &\textbf{0.9423} &0.9383 &0.5386 &0.6845 &0.9821 &0.9499 &0.8140\\
     FCVG &0.9285 &0.9347 &0.5366 &0.6682 &0.9856 &0.9656 &0.8166\\
     CogVX-FT &0.9295 &0.9419 &0.5492 &0.6856 &0.9851 &0.9618 &0.8214\\
     CogVX-BD &0.8718 &0.9232 &0.5137 &0.6319 &0.9800 &0.9727 &0.7931\\
     \hline
     \textbf{Ours} &0.9368 &\textbf{0.9494} &\textbf{0.5527} &\textbf{0.6982} &0.9919 &0.9757 &\textbf{0.8373}\\
     \hline
   \end{tabular}
   }
   \caption{\textbf{VBench dimensional scores of our EF-VI and other baselines on the Pexels dataset.}}
   \label{Vbench on Pexels}
 \end{table*}
\appendix

\section{Harm of Temporal Reverse}
When training, given an $F$-frame video $\boldsymbol{x} \in \mathbb{R}^{F\times 3\times H\times W}$, I2V-DM first uses a variational autoencoder~(VAE)~\cite{kingmaauto, van2017neural} to compress $\boldsymbol{x}$ into the latent representation $\boldsymbol{z} \in \mathbb{R}^{f\times c\times h\times w}$, which can be expressed as:
\begin{equation}
\label{eq14}
    \boldsymbol{z}=\mathcal{E}(\boldsymbol{x}),
\end{equation}
\begin{equation}
\label{eq15}
    \boldsymbol{x}\approx \mathcal{D}(\boldsymbol{z}),
\end{equation}
where $\mathcal{E}$ and $\mathcal{D}$ represent the encoder and the decoder of the VAE, respectively.
Since the reconstruction by the decoder $\mathcal{D}$ has a distortion, we use $\approx$ in Eq. \ref{eq15}.
For U-net-based I2V-DM, such as SVD\cite{blattmann2023stable}, this compression is conducted only in the spatial dimension, which satisfies:
\begin{equation}
    \textit{Flip}(\boldsymbol{z})=\mathcal{E}(\textit{Flip}(\boldsymbol{x}))
\end{equation}
\begin{equation}
    \textit{Flip}(\boldsymbol{x}) \approx \mathcal{D}(\textit{Flip}(\boldsymbol{z}))
\end{equation}
Thus, for U-net-based I2V-DM, $\textit{Flip}(\boldsymbol{z})$ only reverses the temporal order of video frames, while preserving its representation content.
However, for recent and more powerful transformer-based I2V-DM models, such as CogVideoX-5B-I2V~\cite{yang2025cogvideox} and Wan2.1~\cite{wang2025wan}, in addition to spatial compression, their VAEs utilize causal 3D structures to perform causal temporal compression for videos.
In this situation, the above equation breaks:
\begin{equation}
    \textit{Flip}(\boldsymbol{z}) \neq \mathcal{E}(\textit{Flip}(\boldsymbol{x})) 
\end{equation}
\begin{equation}
    \textit{Flip}(\boldsymbol{x}) \neq \mathcal{D}(\textit{Flip}(\boldsymbol{z})) 
\end{equation}
Furthermore, $\textit{Flip}(\cdot)$ disrupts the forward causal relationship of $\boldsymbol{z}$, leading $\textit{Flip}(\boldsymbol{z})$ to fall outside the representation space of the VAE (the input representation of the following denoiser).
We visually display this by utilizing SVD VAE, CogVideoX-5B-I2V VAE, and Wan2.1 VAE to encode the video, flip the compressed latent representation, and reconstruct the video based on the reverse latent representation.
As shown in Fig.~\ref{vae}, the SVD VAE can well reconstruct the temporally-reversed video from the temporally-reversed latent representation.
However, the reconstruction results of the other two VAEs are poor, indicating that $\textit{Flip}(\boldsymbol{z})$ is out of their representation space.

\section{Calculation Details of Evaluation Metrics}
In practice, the frame numbers of the videos generated by different methods vary.
Therefore, to fairly conduct quantitative comparisons between our EF-VI and other baselines, we follow most related works~\cite{zhu2024generative, wang2024generative, yang2025vibidsampler}, which curate ground truth video clips containing 25 frames.
For LPIPS, when the frame number of videos generated by a method differs from the ground truth (25 frames), we first uniformly extract 25 frames from those generated videos before calculation.
For FID and FVD, following \cite{ma2024latte}, we uniformly extract 16 frames from both the ground truth and generated video clips for calculation.
For VBench, following \cite{zhang2025motion}, we evaluate the generated videos across six dimensions: subject consistency, background consistency, aesthetic quality, imaging quality, motion smoothness, and temporal flickering.
In Tab.~\ref{Vbench on DAVIS} and Tab.~\ref{Vbench on Pexels}, we provide the detailed dimensional scores of our EF-VI and other baselines on the DAVIS and Pexels datasets, respectively, which serve as the basis for the VBench scores in Tab.~\ref{quantitative comparison}.
After calculating the dimensional scores, we obtain a final weighted average score using the officially provided weights~\cite{huang2024vbench}, expressed as:
\begin{equation}
    s_{f}=\sum_{j=1}^{6} w_{j} \cdot \frac{s_{j}-s_{j}^{\text{max}}}{s_{j}^{\text{min}}-s_{j}^{\text{max}}}.
\end{equation}
Here, $s_f$ is the final score, $s_j (1\leq j\leq6)$ are the dimensional scores, $w_j$ is the weight for the $j$-th dimension, and $s_{j}^{\text{max}}$, $s_{j}^{\text{min}}$ are also the officially provided dimensional normalization factors.

\section{Utilizing Vision Language Model for Video Inbetweening}
\begin{table}[t]
   \centering
   \scalebox{1.0}{
   \begin{tabular}{ccccc}
     \hline
     Prompt Types &LPIPS$\downarrow$ &FID$\downarrow$ &FVD$\downarrow$ &VBench$\uparrow$\\
     \hline
     D-P &0.2266 &27.26 &567.54 &0.8129\\
     U-P &0.2168 &23.63 &477.87 &0.8169\\
     VLM-P &0.2096 &\textbf{22.30} &382.03 &\textbf{0.8240} \\
     VLM-R-P &\textbf{0.2064} &22.95 &\textbf{380.12} &0.8233 \\
     \hline
   \end{tabular}
   }
   \caption{\textbf{Quantitative comparison of different types of text prompts.}}
   \label{vlm_quantitative_comparison}
 \end{table}
 \begin{figure*}[t]
    \centering
    \includegraphics[width=1.0\linewidth]{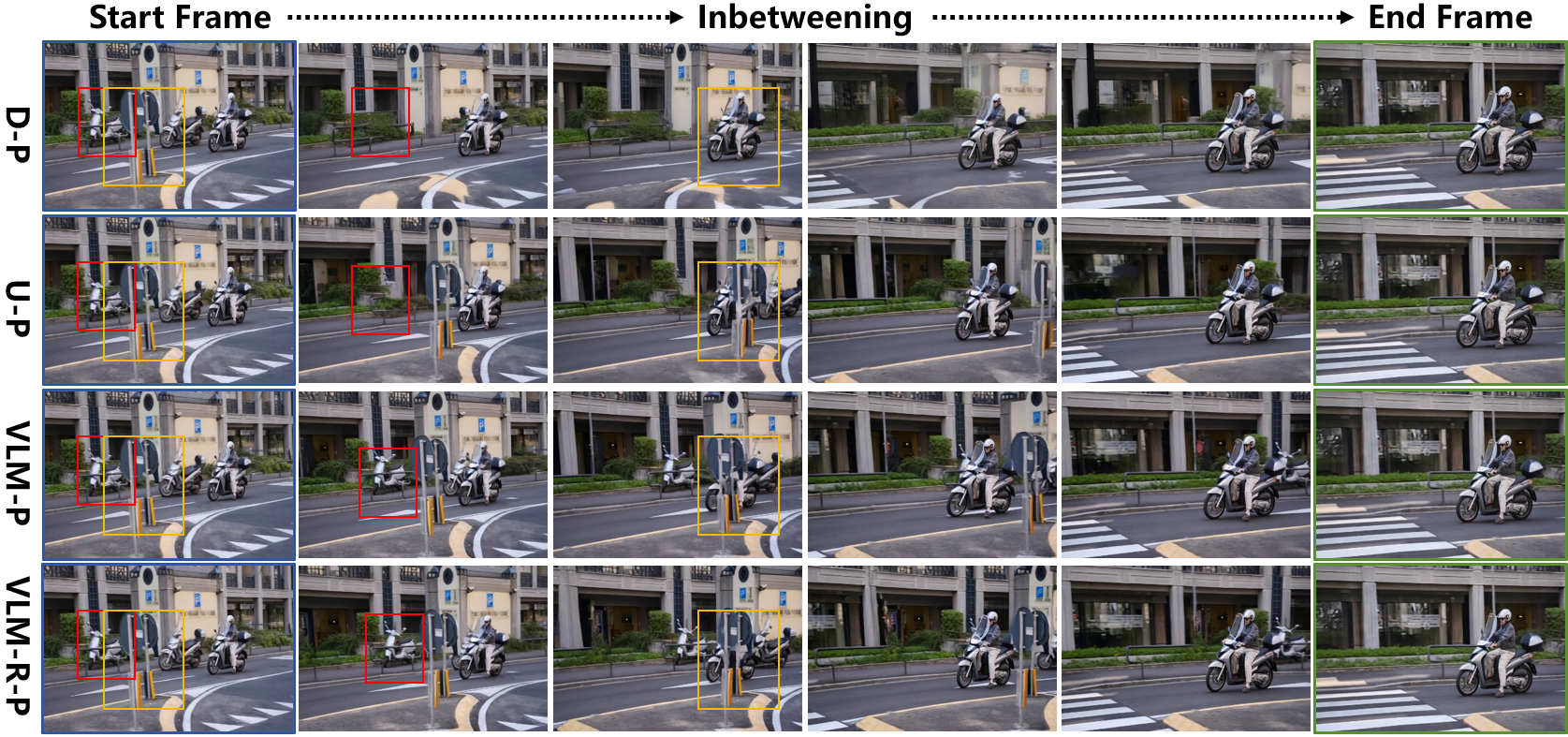}
    \caption{\textbf{Visual comparison between different types of text prompts.}
Utilizing VLM to predict or refine text prompts can enhance the performance of our EF-VI.}
\label{vlm_visual_result}
\end{figure*}
Since the intermediate content generated by some methods and our EF-VI is controlled by the constraints of the start and end frames, as well as the text prompts, for a fair quantitative comparison, we utilize a widely used vision language model~(VLM), Qwen2.5-VL-7B~\cite{Qwen2.5-VL}, to automatically predict the text prompts based solely on the start and end frames.
Specifically, we send each start-end-frame pair and a text instruction, \textit{“This is the first and last frame of a video clip. Describe this video in one continuous paragraph.”}, to Qwen2.5-VL-7B.
It will produce a detailed text prompt for this frame pair.
We conduct an ablation study to explore the effect of the text prompt on the output video quality.
Besides the above VLM-predicted prompt, denoted as \textbf{VLM-P}, we consider three other types of text prompts:
(1) Default prompt, denoted as \textbf{D-P}: We use the prompt, \textit{"This is a clear, smooth, and high-quality video."}, for every input pair.
(2) User-provided short prompt, denoted as \textbf{U-P}: We mimic the habits of actual users who provide a brief text description for an input frame pair, such as \textit{"A man is wearing a grayish-white helmet and riding a motorcycle on the road."}.
(3) LLM-refined prompt, denoted as \textbf{VLM-R-P}: we send each start-end-frame pair and the corresponding \textbf{U-P}, as well as a text instruction, \textit{“This is the first and last frame of a video clip.
U-P is a short description of it.
Please describe this video in one continuous paragraph in detail.”}, to Qwen2.5-VL-7B.
We then utilize the model's output as the final text prompt.
The quantitative results on the DAVIS dataset are provided in Tab. \ref{vlm_quantitative_comparison}.
They demonstrate that utilizing the vision language model to predict long and detailed prompts improves the performance of our method.
Using VLM-P and VLM-R-P achieves significantly better scores on all metrics compared to U-P and D-P.
The visual comparison in Fig.~\ref{vlm_visual_result} further proves the superiority of leveraging VLM.
Without a detailed prompt as a condition, some objects in the video vanish, leading to appearance collapse.
We believe that utilizing VLM to predict or refine text prompts can bring performance gains because prompts handled by VLM are more aligned with the text prompts used in training than the default prompts or user-provided short prompts.

\begin{figure*}[ht]
    \centering
    \includegraphics[width=1.0\linewidth]{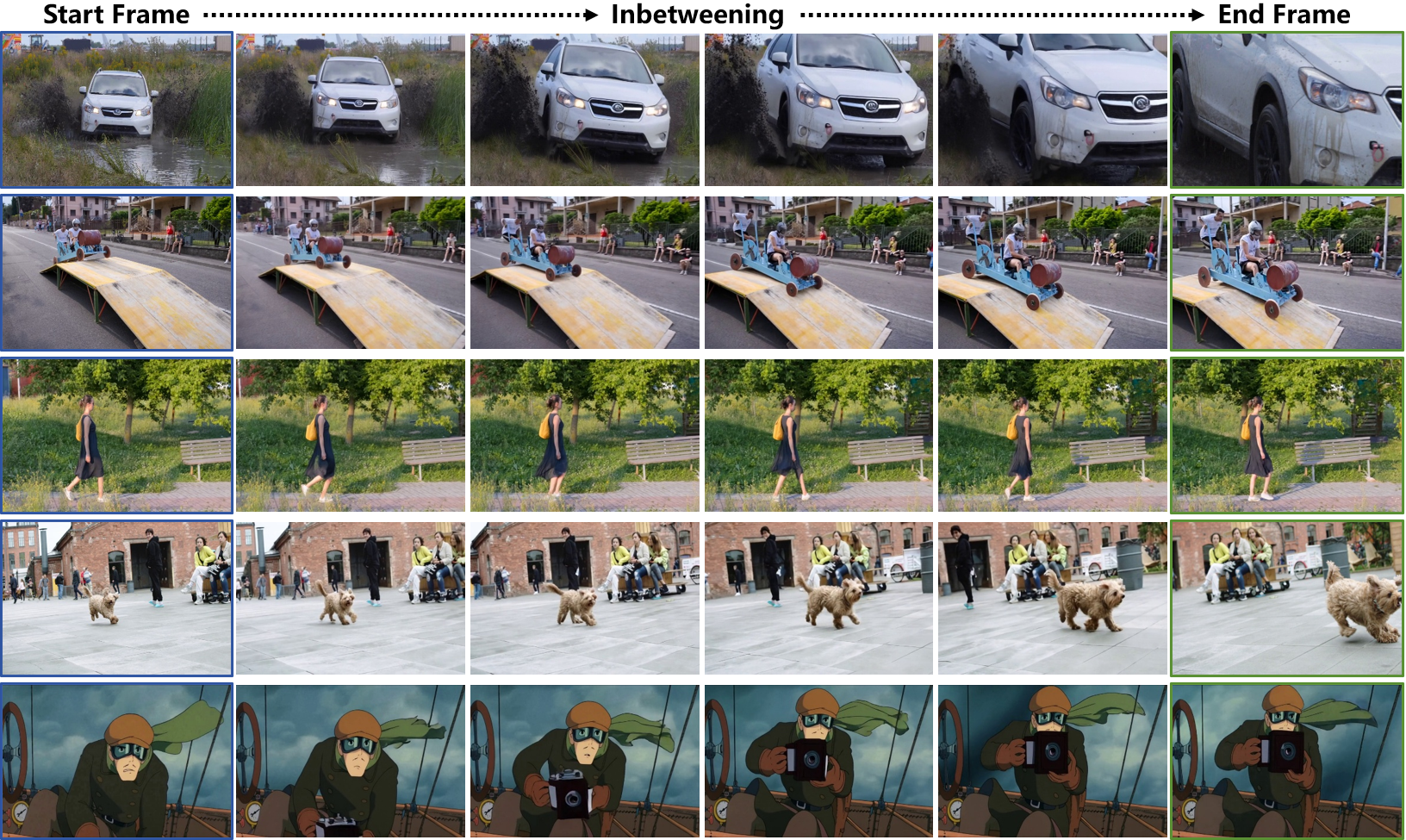}
    \caption{\textbf{Some challenging examples of our EF-VI for video inbetweening.}
Due to the enhanced injection, our EF-VI can generate harmonious intermediate content in complex scenarios involving large and complicated motions of vehicles, people, animals, and cartoon characters, demonstrating the effectiveness of our method.
}
\label{show_case}
\end{figure*}

 \begin{figure*}[t]
    \centering
    \includegraphics[width=1.0\linewidth]{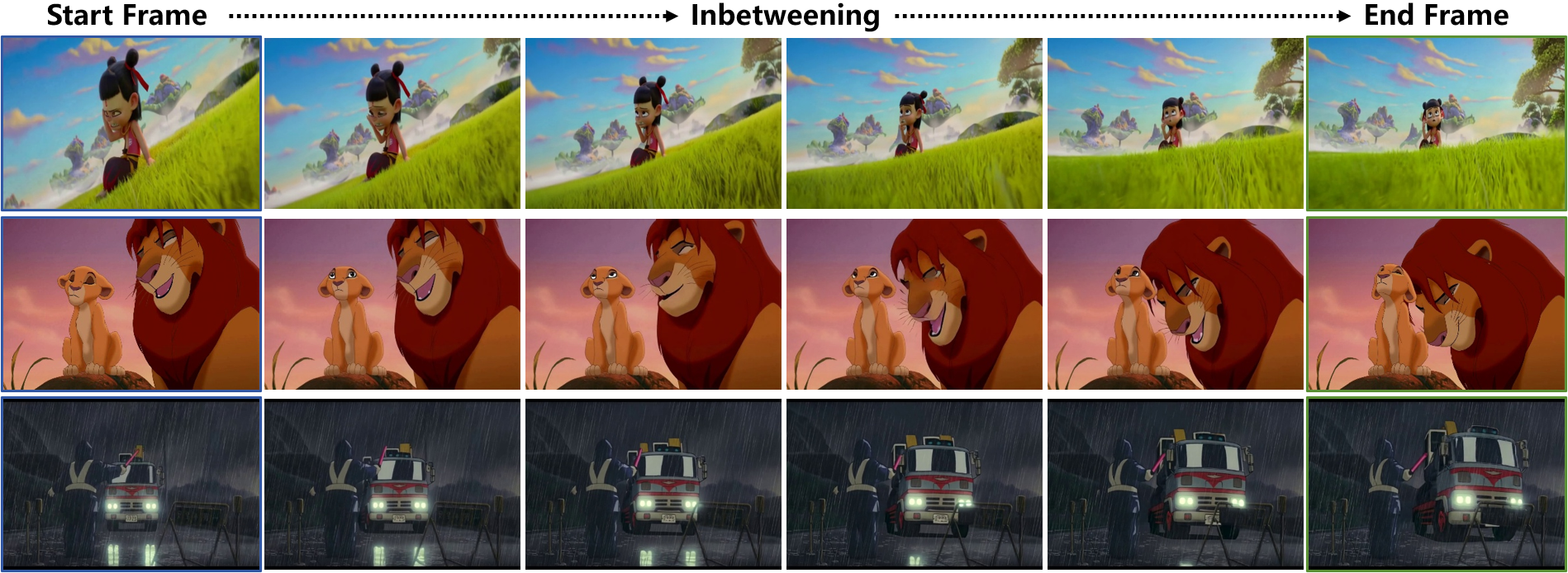}
    \caption{\textbf{Visual results of our EF-VI for cartoon video inbetweening.}
    The generated high-quality cartoon videos demonstrate a strong generalization ability of our method. 
}
\label{cartoon video inbetweening}
\end{figure*}

 \begin{figure*}[t]
    \centering
    \includegraphics[width=1.0\linewidth]{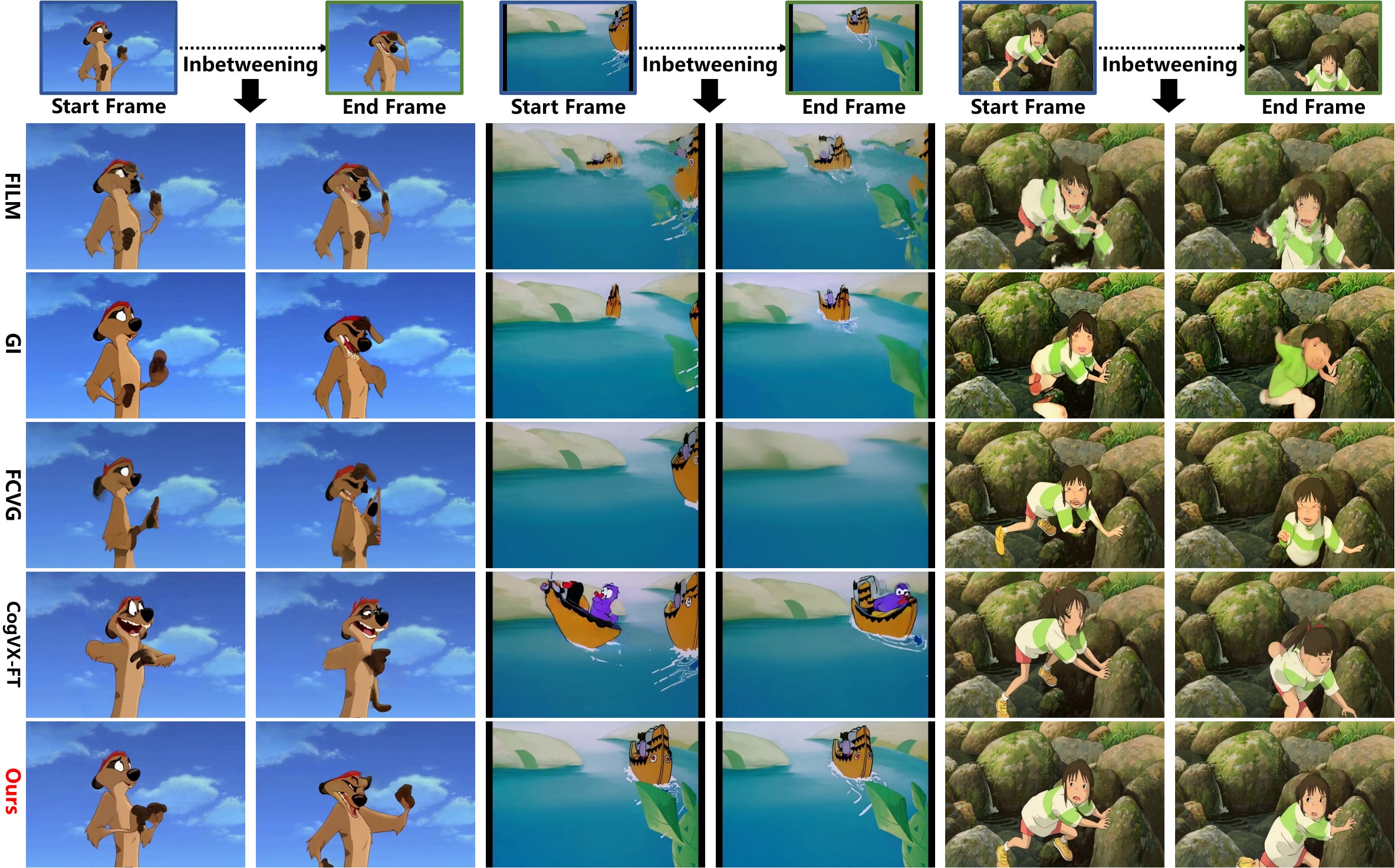}
    \caption{\textbf{Visual comparison between our EF-VI and some advanced methods on cartoon video inbetweening.}
Our EF-VI achieves better visual quality when generalized to cartoon data.}
\label{cartoon visual comparison}
\end{figure*}

\begin{figure*}[t]
    \centering
    \includegraphics[width=1.0\linewidth]{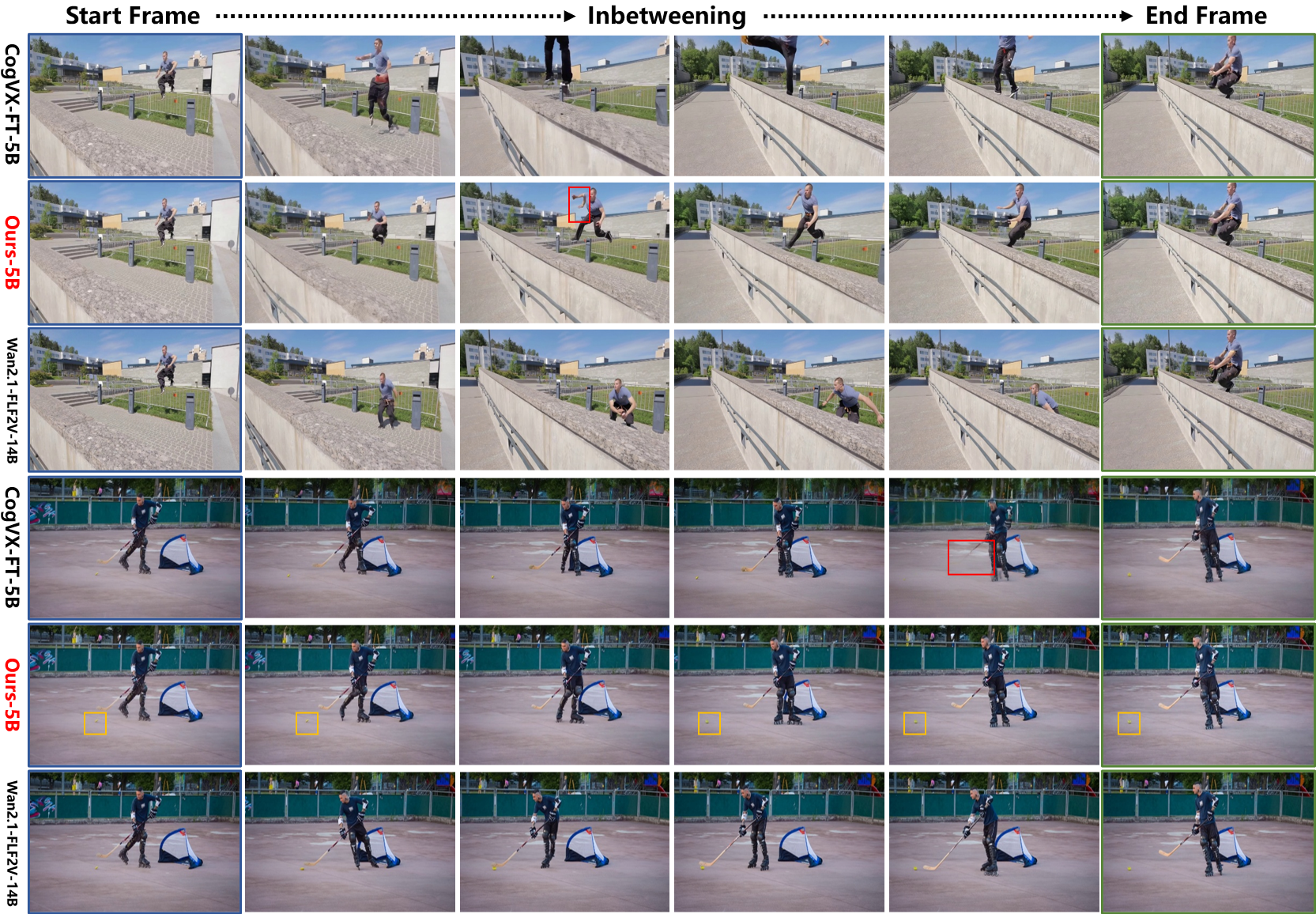}
    \caption{\textbf{Visual comparison of Wan2.1-FLF2V-14B~\cite{wang2025wan}, CogVX-FT~\cite{CogVideoX-FT}, and our EF-VI.}
 Wan2.1-FLF2V-14B improves the quality of generated videos at the cost of much more computation.}
\label{comparison with wan}
\end{figure*}

\begin{table}[t]
   \centering
   \scalebox{1.0}{
   \begin{tabular}{cccc}
     \hline
     \multirow{2}{*}{Methods} &Motion &Content &Overall\\
     &Quality &Fidelity &Attraction \\
     \hline
     FILM &15 &9 &9\\
     GI &81 &72 &67\\
     FCVG &9 &11 &14\\
     CogVX-FT &118 &94 &88\\
     Ours &\textbf{737} &\textbf{774} &\textbf{782} \\
     \hline
   \end{tabular}
   }
   \caption{\textbf{The detailed vote counts of the five methods.}}
   \label{user study details}
 \end{table}
 
\section{More Visual Results of EF-VI}
Due to the enhanced injection, our EF-VI can generate harmonious intermediate content in complex scenarios involving large and complicated motions of vehicles, people, animals, and cartoon characters, demonstrating the effectiveness of our method.

 \section{Details of User Study}

For our user study, the full text of the instructions given to participants is:
\textit{"Hello! Thank you for participating in this survey. We will present 30 groups of video clips.
For each group, the start and end frames are identical (predetermined images), while the intermediate content is generated by five different AI video inbetweening models, respectively. (AI video inbetweening refers to the process where a model generates intermediate transition content based on given initial and final frames).
Your task is to evaluate each group of videos based on the following three dimensions:
\textbf{(1) Motion quality}: Whether the movement of objects and scenes is natural and smooth, without abrupt motion or sudden content jumps.
\textbf{(2) Content fidelity}: Whether the shape and form of objects and scenes match their real-world appearance, without unreasonable deformations or chaotic distortions.
\textbf{(3) Overall attraction}: Comprehensive visual quality and personal preference.
For each dimension, please select the video that you think performs best.
Your feedback is essential to us!"}.
The detailed vote counts are shown in Tab. \ref{user study details}.

\begin{table}[t]
   \centering
   \scalebox{1.0}{
   \begin{tabular}{ccccc}
     \hline
     Methods &LPIPS$\downarrow$ &FID$\downarrow$ &FVD$\downarrow$ &VBench$\uparrow$ \\
     \hline
        FILM &\underline{0.1962} &44.90 &691.57 &\underline{0.8390} \\
     EMA-VFI &0.2503 &80.10 &784.98 &0.7731 \\
     Framer &0.2324 &53.77 &538.81 &0.7997\\
     GI &0.2284 &45.67 &542.42 &0.8308 \\
     FCVG &0.2524 &48.05 &562.63 &0.8260 \\
     CogVX-FT &0.2350 &\underline{40.72} &\underline{466.50} &0.8230 \\
     CogVX-BD &0.3119 &71.4 &597.02 &0.7931 \\
     \hline
     \textbf{Ours} &\textbf{0.1959} &\textbf{37.39} &\textbf{439.63} &\textbf{0.8403} \\
     \hline
   \end{tabular}
   }
   \caption{\textbf{Quantitative Comparison between our EF-VI and some advanced methods on cartoon video inbetweening.}
   The best and second-best scores for each metric are \textbf{bolded} and \underline{underlined}.}
   \label{animate comparison}
 \end{table}
\section{Cartoon Video Inbetweening}
As shown in Fig.~\ref{cartoon video inbetweening}, our EF-VI can still produce visually high-quality results even in challenging scenarios, such as complex actions of cartoon characters or animals, and large movements of vehicles or cameras, demonstrating a strong generalization ability of our method.
For quantitative comparison, we collect 100 cartoon video clips from the internet, including Japanese, American, and Chinese animations.
According to the results shown in Tab.~\ref{animate comparison}, our EF-VI achieves the best scores across all metrics.
For example, our EF-VI's FID and FVD are lower than those of the second-best method, CogVX-FT, which utilizes the same base I2V-DM as ours.
The superiority of our method in cartoon video inbetweening is further reflected in Fig.~\ref{cartoon visual comparison}.
For example, as shown in the middle two columns of Fig.~\ref{cartoon visual comparison}, the dynamics and appearances of the ship in the intermediate frames generated by other methods are chaotic.
On the contrary, our EF-VI achieves much better intermediate transitions with consistent dynamics and appearances.

\section{Failed Cases and Comparison with Wan2.1-FLF2V-14B}
\begin{table}[t] 
   \centering
   \scalebox{0.76}{
   \begin{tabular}{ccccc}
     \toprule
     Methods &Video Size &Step &Time~(s) &Model Size \\ 
     \midrule
     Wan2.1-FLF2V-14B &81 $\times$ 1280 $\times$ 720 &50 &3829.85 &14B \\
     CogVX-FT &49 $\times$ 720 $\times$ 480 &50 &234.89 &5B  \\
     \textbf{Ours} &49 $\times$ 720 $\times$ 480 &50 &237.82 & 5B \\
     
     \bottomrule
   \end{tabular}
   }
   \caption{\textbf{Inference Times of Wan2.1-FLF2V-14B~\cite{wang2025wan}, CogVX-FT~\cite{CogVideoX-FT} and our EF-VI.}
 Due to a larger model size, Wan2.1-FLF2V-14B has a much greater inference consumption.}
   \label{wan effiency}
 \end{table}
Our method's performance is limited by the generation capability of its base model (CogVideoX-5B-I2V).
It struggles to maintain consistent dynamics and appearances in scenarios involving fast or large-scale human movements and the motion of small objects.
As shown in Fig.~\ref{comparison with wan}, the person's body is inconsistent, and the ball suddenly disappears.
A potential approach for improvement is to scale up the model, exemplified by the recently proposed industrial model Wan2.1-FLF2V-14B~\cite{wang2025wan}, which was developed concurrently with our work.
According to the visual comparison in Fig.~\ref{comparison with wan}, the larger model, Wan2.1-FLF2V-14B, yields better visual results and reduces distortion in intermediate content.
However, as shown in Tab. \ref{wan effiency}, scaling up the model will lead to significantly greater inference consumption.
For video inbetweening, designing efficient methods that work well across various scenarios remains challenging and worthy of research in the community.

\section{Instruction of Code and Data}
In the supplementary material, we have zipped the training, inference, and evaluation codes into a file named \textbf{code.zip}.
Additionally, we also provide some examples of our training and testing data, which are placed into a folder named \textbf{data\_examples}.
The complete datasets will be released after the work is accepted.

\section{Societal impacts}
Our proposed EF-VI can efficiently generate high-quality results for video inbetweening.
This novel technology may bring about potential societal implications.
On the one hand, providing more visually harmonious intermediate content benefits fields such as film production and animation creation.
On the other hand, this also raises ethical and safety concerns.
The ease of generating high-quality intermediate content could lead to a surge in the production of harmful or misleading content, such as deepfakes, potentially exacerbating issues of misinformation and privacy invasion.
We condemn the misuse of generative AI to harm individuals or spread misinformation.


\end{document}